%% file: acl_latex.tex
\pdfoutput=1

\documentclass[11pt]{article}

\usepackage[final]{acl}

\usepackage{times}
\usepackage{latexsym}

\usepackage[T1]{fontenc}

\usepackage[utf8]{inputenc}

\usepackage{microtype}

\usepackage{inconsolata}

\usepackage{graphicx}

\usepackage{tikz}
\usetikzlibrary{positioning, shapes.geometric, arrows.meta}

\usepackage{multirow}
\usepackage{booktabs}
\usepackage{pifont}
\usepackage{hyperref}

\usepackage{comment}
\usepackage{arydshln}

\usepackage{subcaption}
\usepackage{graphicx}

\usepackage{multicol}
\usepackage{enumitem}
\usepackage{xcolor}

\usepackage[group-separator= \mathrm{,}]{siunitx}

\title{Meaningful Pose-Based Sign Language Evaluation}

\author{
\parbox{0.8\linewidth}{\centering
Zifan Jiang$^{1}$, Colin Leong*$^{2}$, Amit Moryossef*$^{1, 3}$, \\
Oliver Cory$^{4}$, Maksym Ivashechkin$^{4}$, Neha Tarigopula$^{5, 6}$, Biao Zhang$^{7}$, \\
Anne G\"ohring$^{1}$, Annette Rios$^{1}$, Rico Sennrich$^{1}$, Sarah Ebling$^{1}$}
\\
$^1$University of Zurich
$^2$University of Dayton
$^3$\href{https://sign.mt}{sign.mt}
\\
$^4$University of Surrey
$^5$Idiap Research Institute
$^6$EPFL
$^7$Google DeepMind
\\
\texttt{jiang@cl.uzh.ch}
}

\begin{document}
\maketitle
\def\thefootnote{*}\footnotetext{Equally contributed as co-second authors.}
\def\thefootnote{\arabic{footnote}}
\begin{abstract}
We present a comprehensive study on meaningfully evaluating sign language utterances in the form of human skeletal poses. 
The study covers keypoint distance-based, embedding-based, and back-translation-based metrics. 
We show tradeoffs between different metrics in different scenarios through (1) automatic meta-evaluation of sign-level retrieval, and (2) a human correlation study of text-to-pose translation across different sign languages. 
Our findings, along with the open-source \href{https://github.com/sign-language-processing/pose-evaluation}{\texttt{pose-evaluation}} toolkit, provide a practical and reproducible approach for developing and evaluating sign language translation or generation systems.
\end{abstract}

\section{Introduction}

Automatic evaluation metrics are essential for assessing the quality of automatically generated language content and tracking progress over time. 
For instance, machine translation (MT) studies rely heavily on BLEU \cite{papineni-etal-2002-bleu}, even though newer metrics have shown stronger correlation with human judgment \cite{freitag-etal-2022-results}. 
This trend continues in sign language processing (SLP; \citet{bragg2019sign, yin-etal-2021-including}), an interdisciplinary subfield of natural language processing and computer vision.
Sign language translation (SLT; \citet{muller-etal-2022-findings,muller-etal-2023-findings,de2023machine}), denoting the part of SLP concerned with translating sign language videos into spoken language text, reuses text-based metrics.

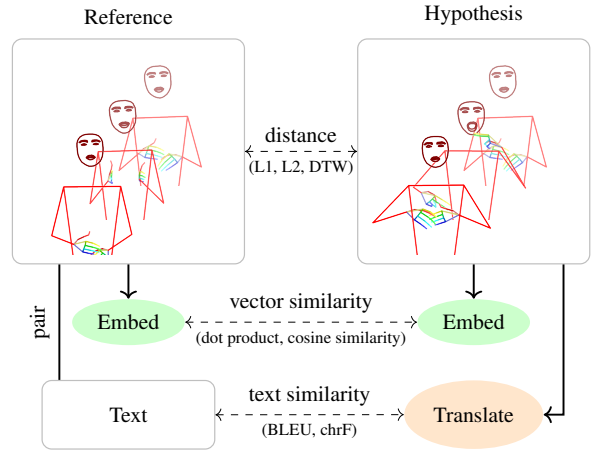
\begin{figure}[ht]
  \centering
  
  \resizebox{\linewidth}{!}{\input{figures/title-figure}}
  \caption{Pose‐based evaluation taxonomy overview. We compare a reference and a hypothesis pose sequence by one of the following three ways: (a) computing distance‐based metrics directly on the keypoint sequences, optionally aligned by dynamic time wrapping (DTW); (b) encoding each sequence into a shared embedding space and measuring similarity; and (c) back‐translating the hypothesis poses into text to apply conventional machine translation metrics on text.}
  \label{fig:title}
\end{figure}

\citet{muller-etal-2023-considerations} puts forward concrete suggestions on evaluating generated text (especially glosses) in a sign language context.
They suggest always computing metrics with standardized tools (e.g., SacreBLEU \cite{post-2018-call} for BLEU) and reporting the metric signatures for reproducibility and fair comparison with other work. 
The opposite direction—generating or translating into sign language utterances (usually from source text)—presents additional challenges for evaluation. Namely, standardized metrics, tooling, and correlation with human evaluation are lacking.

In this work, we systematically examine the metrics employed for evaluating sign language output, especially formatted as human skeletal poses \citep{zheng2020deep} that contain motion of signing (e.g., MediaPipe Holistic; \citet{lugaresi2019Mediapipe, Mediapipe2020holistic}).
We start by a literature review of current research practices in \S\ref{sec:related}, and summarize two major families of metrics: 
(a) distance-based metrics (\S\ref{sec:distance}) informed by human motion generation (\S\ref{sec:motion}) and sign language assessment (SLA; \S\ref{sec:assessment}), assuming the access to reference poses and then computing the distance from the predicted poses to the reference poses, either in the raw 2D/3D keypoint space or an embedding space; 
(b) back-translation-based metrics (\S\ref{sec:bt}) borrowed from MT \cite{zhuo-etal-2023-rethinking} and speech translation \cite{zhang-etal-2023-dub}, assuming the pre-existence of a pose-to-text translation model.

After the initial conceptual review, we select, implement, and meta-evaluate typical metrics  along with additional innovative ones proposed by us (as summarised in Figure \ref{fig:title}), through two empirical approaches:
automatic meta-evaluation with a sign-level retrieval task (\S\ref{sec:automatic});
and a sentence-level correlation study between metrics of interest versus deaf evaluator ratings on three text-to-pose MT systems in three spoken-sign language pairs (\S\ref{sec:correlation}).

We find that keypoint distance-based metrics, when carefully tuned, can rival more advanced approaches for sign retrieval and human‐judgment correlation.
On the other hand, embedding-based metrics, including those borrowed from SLA, excel in their own domain but struggle at the sentence level across different systems. 
Back‐translation likelihood emerges as the most consistent metric, highlighting the need for open, standardized pose-to-text models alongside human evaluation.

The source code of the suggested evaluation metrics and the proposed meta-evaluation protocols in \S\ref{sec:automatic} are openly maintained in \href{https://github.com/sign-language-processing/pose-evaluation}{\texttt{pose-evaluation}}, a public GitHub repository.
The human correlation data and evaluation scripts in \S\ref{sec:correlation} are also released in a separate \href{https://github.com/ZurichNLP/text2pose-human-eval}{\texttt{text2pose-human-eval}} repository to encourage future research. 

\section{Related Work}
\label{sec:related}

We discuss four related fields in this section with a special emphasis on the evaluation methodology, and outline recent work in sign language generation (SLG; \S\ref{sec:slg}) in Table \ref{tab:literature_review}. 
The remaining three fields provide additional background relevant to evaluating these SLG systems.

\begin{table*}[!ht]
    \centering
    \resizebox{\linewidth}{!}{%
    \begin{tabular}{l l c c c cc ccccc l}
      \toprule
      \multicolumn{1}{c}{\textbf{Work}} & \textbf{Datasets} & \textbf{Sources} & \textbf{Target} & \multicolumn{2}{c}{\textbf{Model}} & \multicolumn{6}{c}{\textbf{Evaluation Metrics}} \\
      \cmidrule(lr){2-2}
      \cmidrule(lr){3-3}
      \cmidrule(lr){4-4}
      \cmidrule(lr){5-6}
      \cmidrule(lr){7-13}
      & (P,H2, etc.) & (T,G,H) & (M,O,S) & \textbf{</>} & \textbf{$\theta$} & \textbf{D} & \multicolumn{1}{c}{\textbf{</>}} & \textbf{B4} & \textbf{</>} & \multicolumn{1}{c}{\textbf{$\theta$}} & \textbf{Other} \\

      \midrule

      \citet{arkushin2023ham2pose} & DGS Corpus, 3 others & H & O & \href{https://github.com/rotem-shalev/Ham2Pose}{\ding{52}} & \href{https://github.com/rotem-shalev/Ham2Pose/tree/main/models/ham2pose/checkpoints}{\ding{52}} & \ding{52} & \href{https://github.com/rotem-shalev/Ham2Pose/blob/main/metrics.py}{\ding{52}} & n/a & n/a & n/a & - \\

      \midrule

      \citet{stoll2018sign,stoll2020text2sign} & P & G & O & - & n/a & - & - & - & - & - &  SSIM, PSNR, MSE (pixel-wise) \\
      
      \citet{moryossef-etal-2023-open} & Signsuisse & G & M & \href{https://github.com/ZurichNLP/spoken-to-signed-translation}{\ding{52}} & n/a & - & - & - & - & - & - &  \\

      \citet{zuo2024simple} & P,CSL-Daily & G & S & \href{https://github.com/FangyunWei/SLRT/tree/main/Spoken2Sign}{\ding{52}} & n/a & \ding{52} & - & \ding{52} & \href{https://github.com/FangyunWei/SLRT/tree/main/TwoStreamNetwork}{\ding{52}} & \href{https://github.com/FangyunWei/SLRT/tree/main/TwoStreamNetwork\#performance}{\ding{52}} & Frame temporal consistency \\
      
      \midrule
      \citet{saunders2020progressive,saunders2020adversarial,saunders2021continuous,saunders2021mixed} & P & T & O & \href{https://github.com/BenSaunders27/ProgressiveTransformersSLP}{\ding{52}} & - & - & - & \ding{52} & \href{https://github.com/neccam/slt}{\ding{52}} & - & - \\

      \citet{hwang2021non,hwang2023autoregressive} & P,H2 & T & O & \href{https://github.com/Eddie-Hwang/NSLP-G}{\ding{52}} & - & \ding{52} & - & \ding{52} & - & - & Fréchet Gesture Distance \\

      \citet{yin-etal-2024-t2s} & P & T & S & - & - & \ding{52} & - & \ding{52} & - & - & - \\
      \citet{fang2024signdiffdiffusionmodelsamerican,fang2024signllm} & P,H2,4 others & T & O & - & - & \ding{52} & - & \ding{52} & - & - & SSIM, Hand SSIM, FID, etc. \\
      
      \citet{yu2024signavatars} & P,H2,4 others & T,G,H & S & \href{https://github.com/ZhengdiYu/SignAvatars}{\ding{52}} & - & \ding{52} & - & - & - & - & FID, Diversity, MM-Dist, etc. \\
       
      \citet{Baltatzis_2024_CVPR} & H2 & T & S & - & - & \ding{52} & - & \ding{52} & - & - & FID \\

      \citet{zuo2024soke} & P,H2,CSL-Daily & T & S & - & - & \ding{52} & - & \ding{52} & - & - & Latency \\
      
      \bottomrule
    \end{tabular}
    }
    \caption{Literature review of recent works on pose-based sign language generation (May 2025). P=\texttt{RWTH-PHOENIX-Weather2014T}, H2=\texttt{How2Sign}; T=\texttt{Text}, G=\texttt{Gloss}, H=\texttt{HamNoSys}; M=\texttt{MediaPipe}, O=\texttt{OpenPose}, S=\texttt{SMPL-X}; D=\texttt{DTW-MJE} (and other distance-based metrics), B4=\texttt{BLEU-4} (and other back translation-based metrics); </> and $\theta$ represent the availability of source code and model weights for the generation model and the evaluation metrics (including the back translation model if involved), respectively. The check mark symbols (\ding{52}) are clickable links in these columns, and \textit{n/a} denotes not-applicable cases, such as model weights for gloss-based systems and back-translation for HamNoSys input. Other image-based metrics are left as less relevant.}
    \label{tab:literature_review}
\end{table*}

\subsection{Sign Language Understanding}

Sign language recognition \cite{adaloglou2021comprehensive} and translation \cite{de2023machine} are the two most prevalent tasks of understanding sign language from video recordings.
The former aims to classify signing into a fixed vocabulary of signs in a particular sign language, either from isolated video clips of single signs (isolated sign language recognition, ISLR) or continuous video footage spanning multiple signs (continuous sign language recognition, CSLR).
Given its classification nature, the evaluation efficiently utilizes classic statistical metrics, such as accuracy, $F_1$ score, and word error rate.

Early SLT attempts rely on glosses \cite{moryossef-etal-2021-data,muller-etal-2023-considerations}, produced manually by humans or a CSLR model.
\citet{camgoz2018neural,camgoz2020sign} starts end-to-end neural SLT and leads a wave of gloss-free SLT work \cite{zhou2023gloss, zhang2024scaling}, where evaluation is typically done with BLEU and BLEURT 
\cite{sellam-etal-2020-bleurt} but not possible with source-based metrics like COMET and quality estimation models like COMET-QE \cite{chimoto-bassett-2022-comet} due to the input modality constraint on sign language.
WMT-SLT campaigns for two consecutive years \cite{muller-etal-2022-findings,muller-etal-2023-findings} carry out a rigorous human evaluation process as seen in traditional MT research. 
Yet the correlation between automatic evaluation metrics and human judgments in SLT has not been reported; quantifying this correlation would yield valuable insights.

\subsection{Sign Language Generation} 
\label{sec:slg}

The landscape of SLG is more complicated than SLT, with various inputs, namely, 
(a) spoken language \textit{text}; 
(b) sign language \textit{glosses}; 
(c) iconic \textit{phonetic writing systems} of sign language;
(d) textual \textit{phonetic descriptions} of signing,
and various outputs, usually,
2D/3D pose; or
RGB video frames\footnote{For further details about these representations, please refer to the explanatory figures on \href{https://research.sign.mt/}{https://research.sign.mt/}.}.
We note that in the case of (a) \textit{text}, the generation process involves translation from a spoken language to a sign language with possibly reordering and rephrasing of words, while starting with (b), (c), or (d)  merely convert sign language approximated in textual forms into visuals (also known as sign language production\footnote{The terms are sometimes used interchangeably and thus confuse. This work adheres to the broad term of sign language generation, which involves generating signing from any source.}), possibly with a preceding step in the pipeline that translates from (a) \textit{text} to (b) \textit{SignWriting} \cite{jiang-etal-2023-machine}, (c) \textit{glosses} \cite{zhu-etal-2023-neural}, or (d) \textit{descriptions}\footnote{For example, signing HELLO in ASL: dominant B-hand at forehead → short outward stroke; friendly/smiling face.}.
Our work evaluates poses as the primary representation of sign language motion and semantics, deliberately excluding RGB videos to avoid confounding factors such as visual appeal or signer identity. 
Evaluating videos using the same methods is possible after first estimating them into poses.

We present prominent pose-based SLG studies from recent years, along with their evaluation methods, in Table \ref{tab:literature_review}, grouped by input modalities.
Following a similar roadmap as SLT, SLG takes off with a gloss-based cascading approach (text-to-gloss-to-sign; \citet{stoll2018sign,stoll2020text2sign}) and then gradually switches to an end-to-end fashion in a series of follow-up work \cite{saunders2020progressive,saunders2020adversarial,saunders2021continuous,saunders2021mixed}.
Attempts have also been made with alternative phonetic inputs such as HamNoSys \cite{writing:prillwitz1990hamburg,arkushin2023ham2pose}.

Unlike SLT, however, gloss-based baseline approaches for SLG remain competitive and practical choices \cite{moryossef-etal-2023-open,zuo2024simple} due to accessible sign language dictionary resources that enable straightforward mapping of glosses to sign language pose sequences. 
Modern end-to-end approaches utilise vector quantization, 
diffusion models,
and LLMs, 
and the output pose format spans from classic 2D standards such as MediaPipe Holistic and Openpose \citep{openpose} to 3D SMPL-X \cite{Pavlakos_2019_CVPR}.

Popular datasets used in this line of work include RWTH-PHOENIX-Weather 2014T, in German Sign Language (DGS), introduced by \citet{forster-etal-2014-extensions, camgoz2018neural}; CSL-Daily, in Chinese Sign Language (CSL), introduced by \citet{zhou2021improving}; and How2Sign, in American Sign Language (ASL), introduced by \citep{Duarte_CVPR2021}.
We choose Signsuisse \cite{muller-etal-2023-findings} in this work (\S\ref{sec:correlation}) for its multilingual nature and richer vocabulary than others\footnote{PHOENIX and CSL-Daily feature 1066 and 2000 signs.}.

As for evaluation, the \href{https://www.codabench.org/competitions/4854/#/pages-tab}{SLRTP Sign Language Production Challenge 2025} summarises the most common evaluation metrics: 
(a) keypoint distance-based, such as DTW-MJE (Dynamic Time Warping - Mean Joint Error); and
(b) back-translation-based, such as BLEU and BLEURT.
Human evaluation is conducted briefly in \citet{saunders2021continuous,saunders2021mixed}, \citet{Baltatzis_2024_CVPR}, and \citet{zuo2024soke} and more extensively in another concluded campaign--\href{http://cips-cl.org/static/CCL2024/en/cclEval/taskEvaluation/index.html\#Task\%2010:\%20Quality\%20Evaluation\%20of\%20Sign\%20Language\%20Avatars\%20Translation}{Quality Evaluation of Sign Language Avatars Translation} \cite{yuan-etal-2024-translation}.
Unfortunately, like in SLT, the correlation between automatic metrics and human judgments has never been formally validated.
Upon reviewing Table \ref{tab:literature_review}, we spot two significant issues in the current development of SLG:
(a) Most systems and their evaluations are non-reproducible due to the lack of source code and model weights (including the back-translation models if involved).
(b) Cross-work comparisons are unrealistic given the fragmented implementation of the evaluation metrics (in contrast to MT, where standardized tools like SacreBLEU are available).







\subsection{Human Motion Generation}
\label{sec:motion}

Motion generation from natural language is a related field where human pose sequences are synthesized to reflect described actions \cite{tevet2022motionclip, zhang2024motiondiffuse}. 
Evaluation typically involves distance-based metrics (e.g., joint or velocity error), perceptual similarity (e.g., Fréchet Inception Distance adapted to motion), and alignment metrics, such as R-Precision, to measure text-motion coherence. 
However, \citet{voas2023best} shows that many of these automated metrics correlate poorly with human judgment on a per-sample basis. They propose MoBERT, a BERT-based learned evaluator, which achieves higher agreement with human ratings, highlighting the ongoing challenge of designing semantically meaningful motion evaluation.

\subsection{Sign Language Assessment}
\label{sec:assessment}



SLA research compares student-produced signing against canonical references. \citet{cory2024modelling} evaluates sign language proficiency by modeling the natural distribution of signing motion across multiple references and demonstrating a strong correlation with human ratings. 
\citet{TarigopulaSGEval2024,tarigopula2025posterior} proposes a posterior-based analysis of skeletal or spatio-temporal features to assess both manual and non-manual signing components, improving alignment with human evaluation as well.

\section{Evaluation Metrics}
\label{sec:metrics}

In this section, we formally define common evaluation metrics mentioned in related work (\S\ref{sec:related}) and implement them, reusing open-source code where available,  to prepare for the upcoming empirical study on pose evaluation in \S\ref{sec:automatic} and  \S\ref{sec:correlation}.

\subsection{Keypoint Distance-Based Metrics}
\label{sec:distance}

We borrow keypoint distance-based metrics from prior work on sign language generation, notably Ham2Pose \cite{arkushin2023ham2pose}. These metrics, e.g., APE (Average Position Error)—initially developed for general pose estimation and motion analysis—quantify geometric similarity using frame-wise errors and alignment strategies. However, they are not designed for sign language and ignore critical linguistic properties such as signer speed variation, hand dominance, and missing keypoints. Moreover, they have not been systematically validated against human judgments in sign language contexts, motivating our extended investigation.

\begin{figure}[!t]
  \centering
  \begin{subfigure}{0.49\linewidth}
    \includegraphics[trim=350 0 180 250,clip,width=\linewidth]{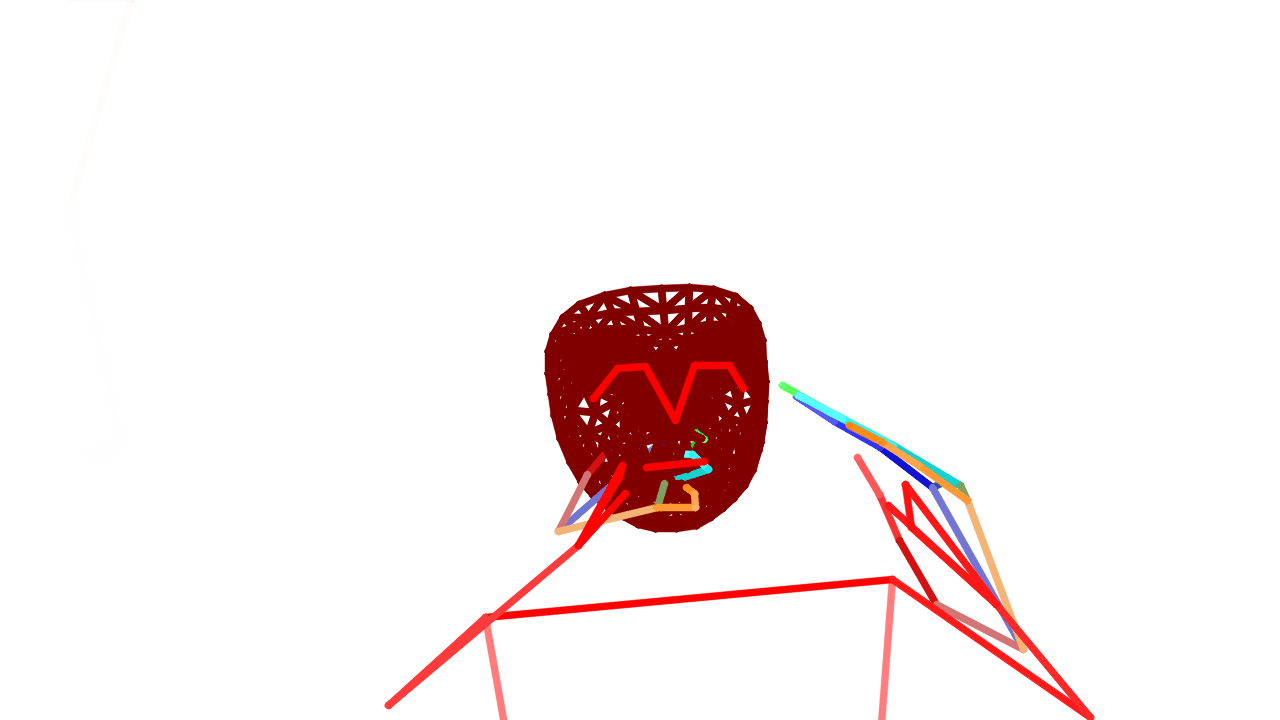}
    \caption{Full (586 Points)\\{\scriptsize \cite{Mediapipe2020holistic}}}
  \end{subfigure}
  \begin{subfigure}{0.49\linewidth}
    \includegraphics[trim=350 0 180 250,clip,width=\linewidth]{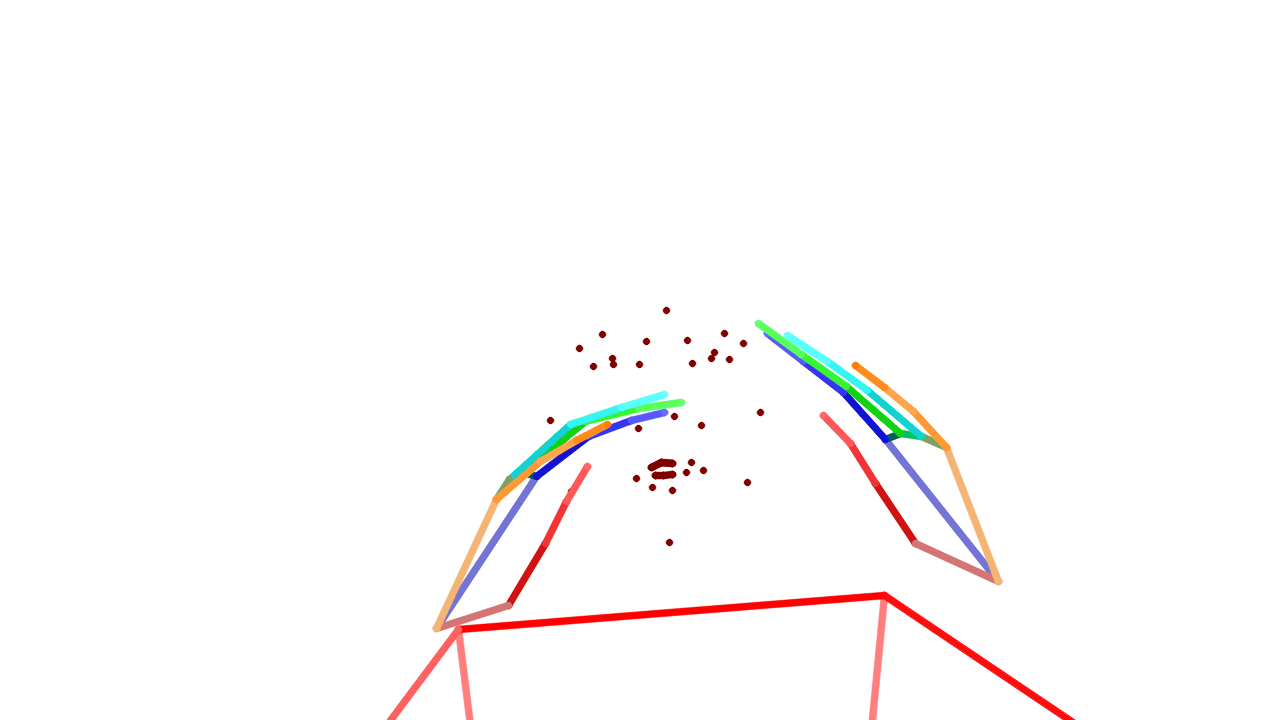}
    \caption{YouTube-ASL (85 Points)\\\cite{uthus_youtube-asl_2023} }
  \end{subfigure}
  \begin{subfigure}{0.49\linewidth}
    \includegraphics[trim=350 0 180 250,clip,width=\linewidth]{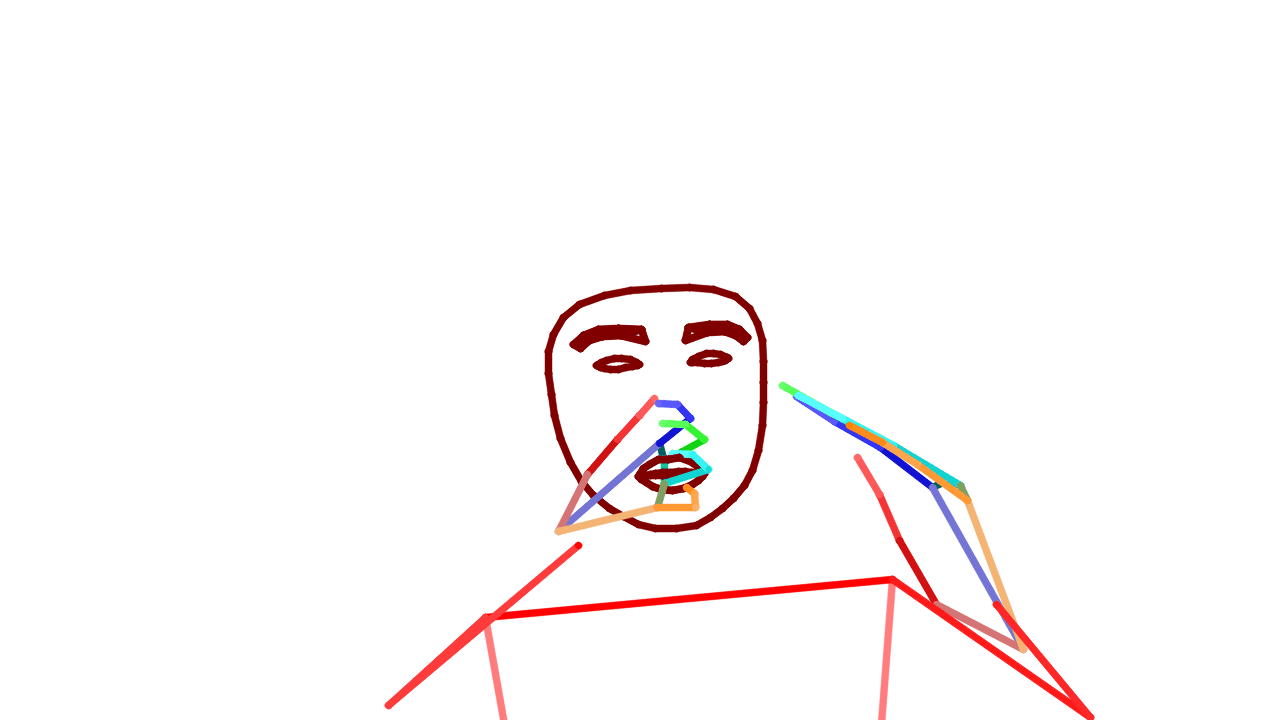}
    \caption{Face-contour (178 Points)\\\cite{moryossef2021pose-format}}
  \end{subfigure}  
  \begin{subfigure}{0.49\linewidth}
    \includegraphics[trim=350 0 180 250,clip,width=\linewidth]{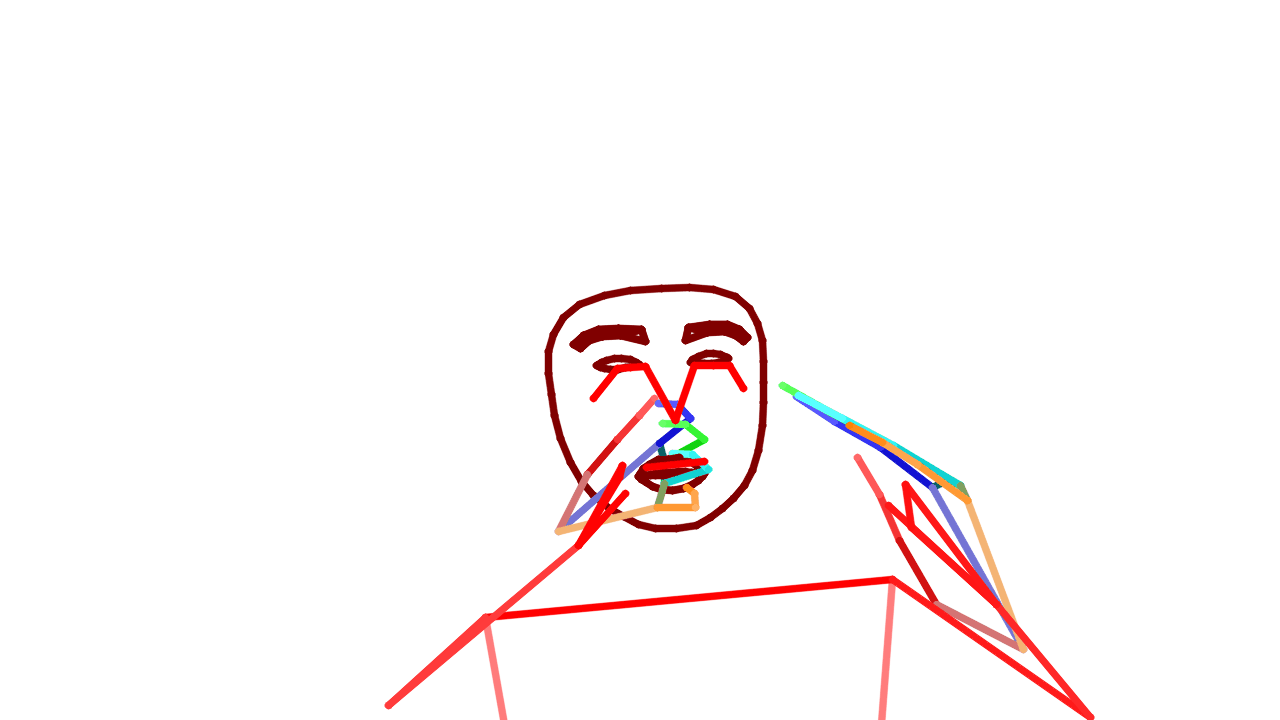}
    \caption{SignCLIP (203 Points)\\\cite{jiang-etal-2024-signclip} }
  \end{subfigure}
  \caption{MediaPipe keypoint selection strategies.}
  \label{fig:keypoint_selection}
\end{figure}

\begin{figure}[!t]
  \centering
  \begin{subfigure}{0.5\linewidth}
    \scalebox{1}[0.5]{\includegraphics[width=\linewidth]{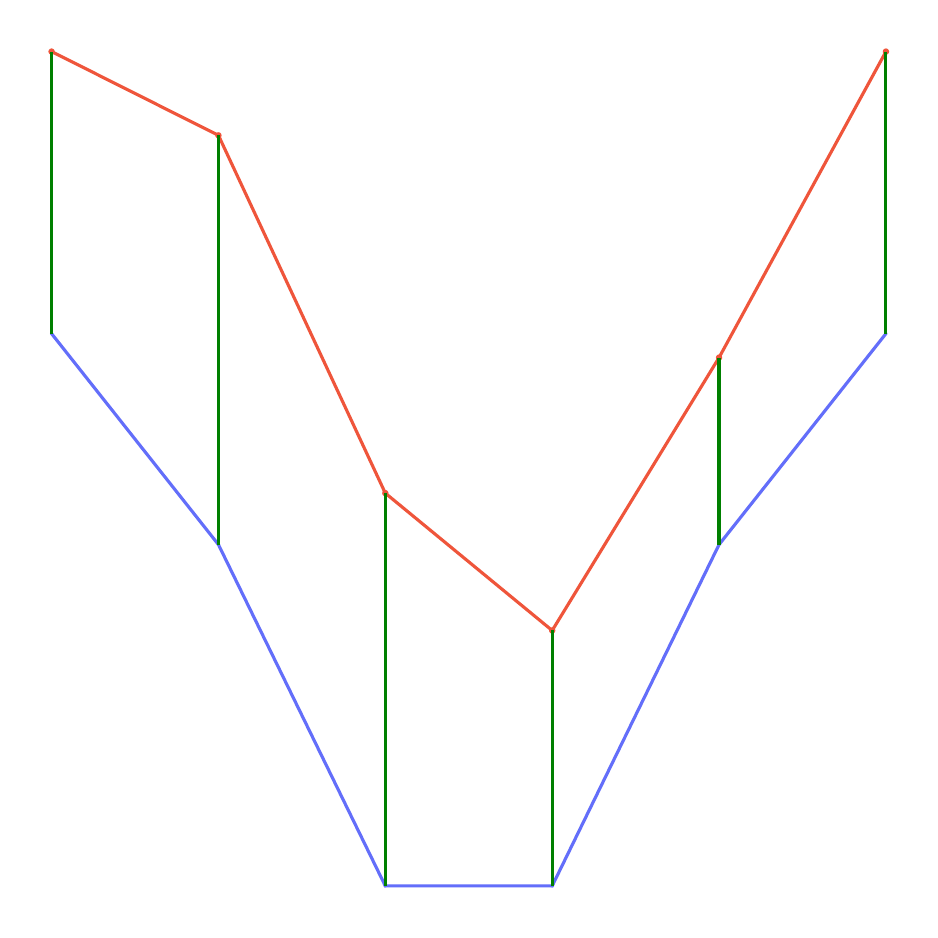}}
    \caption{Equal-Length Sequences}
  \end{subfigure}%
  \begin{subfigure}{0.5\linewidth}
    \scalebox{1}[0.5]{\includegraphics[width=\linewidth]{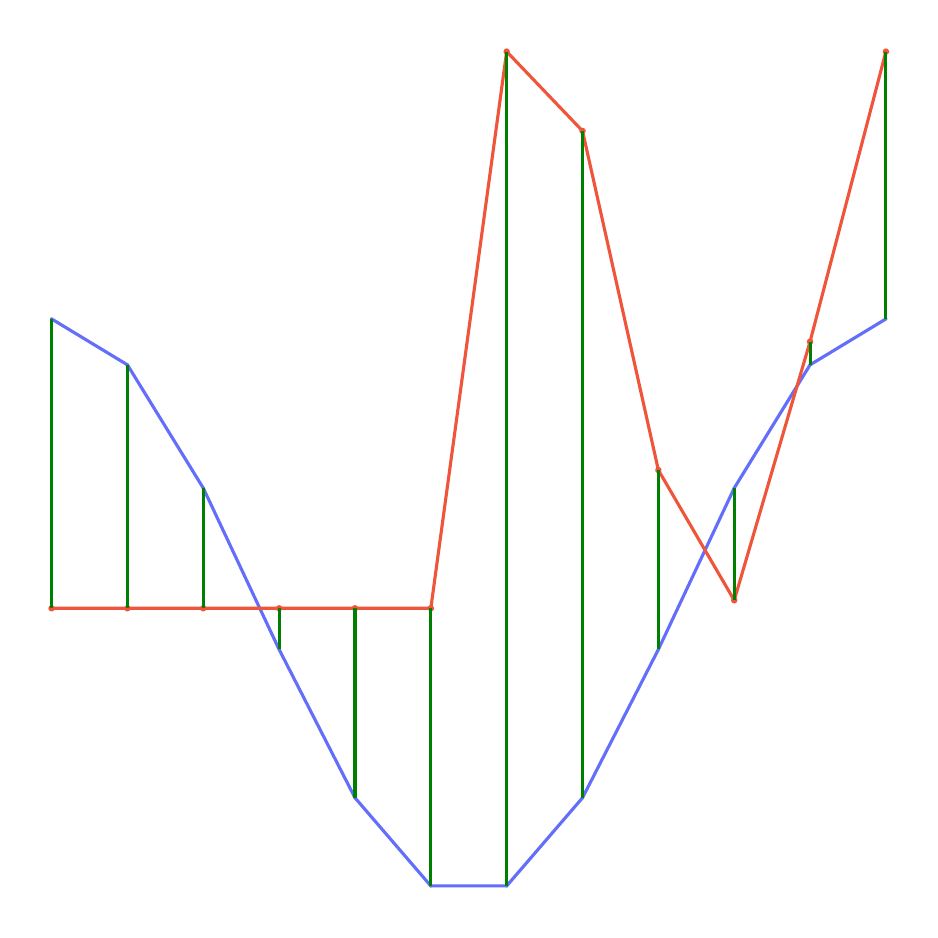}}
    \caption{Zero-Padding}
  \end{subfigure}
  \begin{subfigure}{0.5\linewidth}
    \scalebox{1}[0.5]{\includegraphics[width=\linewidth]{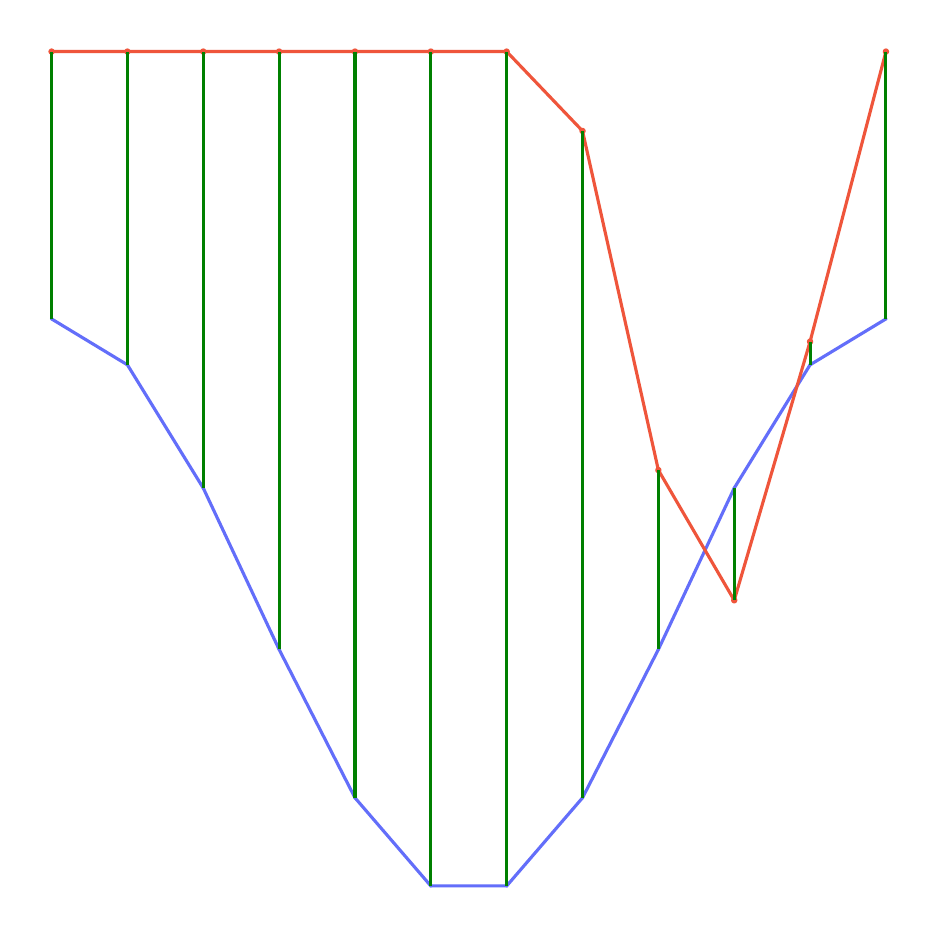}}
    \caption{Pad with First Frame}
  \end{subfigure}%
  \begin{subfigure}{0.5\linewidth}
    \scalebox{1}[0.5]{\includegraphics[width=\linewidth]{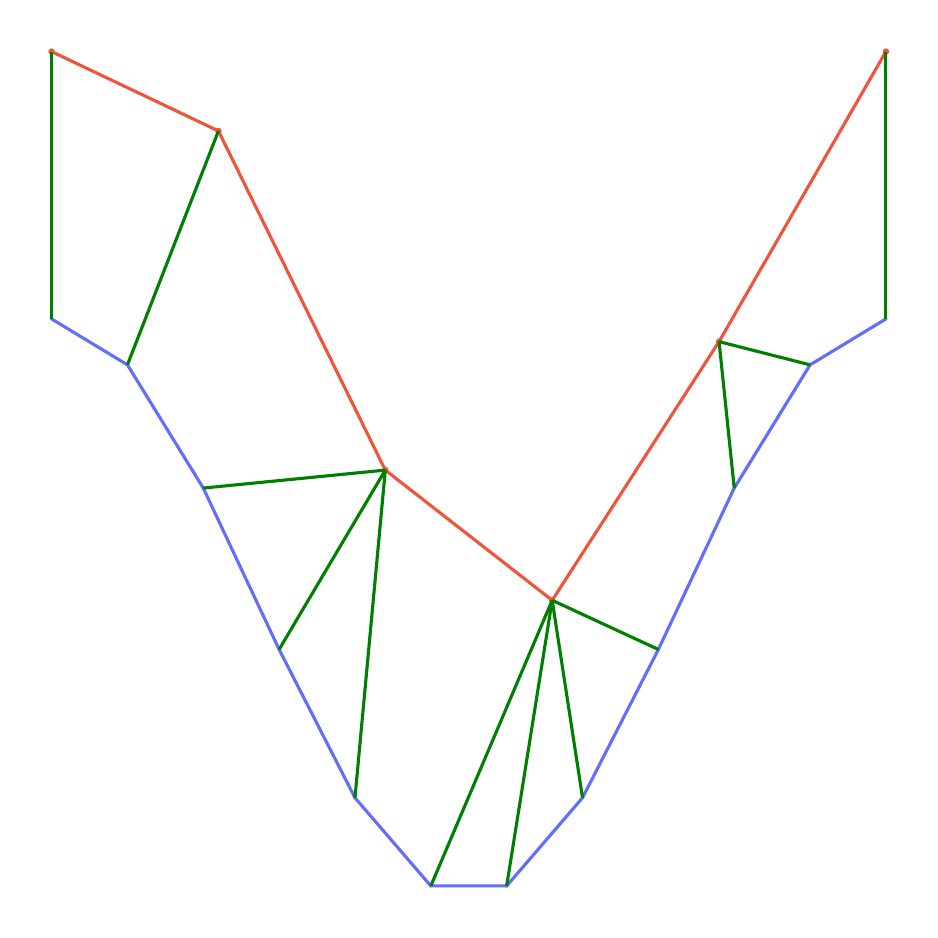}}
    \caption{Dynamic Time Warping}
  \end{subfigure}
  \caption{Sequence alignment (in \textcolor{green}{green}) between a shorter sequence (in \textcolor{red}{red}) and longer sequence (in \textcolor{blue}{blue}). In reality, pose keypoint trajectories are aligned temporally in 3D and then averaged for the whole body. Paddings take values from the first frame or simply 0s.}
  \label{fig:seq_align}
\end{figure}

During (re-)implementation, we identify significant sources of variation that affect the outcomes of distance-based metrics: 
(a) whether and how the coordinate values of the keypoints are normalized (e.g., based on the shoulder position as the origin $(0,0)$ and the shoulder width being 1); (b) whether videos are trimmed to exclude signing-inactive frames; (c) which subset of the keypoints from the pose estimation library is selected for comparison (Figure~\ref{fig:keypoint_selection}; e.g., hands-only vs. full body); (d) how framerate mismatches are handled (e.g., interpolating to a consistent FPS); (e) how masked or missing keypoints are treated (e.g., filled with a value, or a default distance returned, or simply ignored); and (f) how sequences of unequal length are aligned before applying APE (Figure~\ref{fig:seq_align}; e.g., using zero-padding, frame repetition, or DTW). 

We examine these variations and provide a reproducible toolkit that enables tuning these design choices explicitly—including keypoint selection, normalization, and masking, sequence trimming and alignment with different distance measures—rather than inheriting arbitrary defaults.
The toolkit supports the generation of possibly thousands of metric variants to be tested in \S\ref{sec:automatic}.

\subsection{Embedding-Based Metrics}

Rather than operating on the keypoints' raw spatial positions, we categorize embedding-based metrics that calculate distance or similarity in a latent embedding space provided by a trained model.

\subsubsection{Sign Language Assessment Metrics}
\label{sec:sla_matrics}

We adopt two metrics for comparing two poses from the SLA task (\S\ref{sec:assessment}):  the Skeleton Variational Autoencoder (SkeletonVAE) model from \citet{cory2024modelling} and the posterior-based scores from assessment models developed in \citet{TarigopulaSGEval2024}.

\paragraph{SkeletonVAE Score}
The SkeletonVAE is trained to produce a per-frame latent embedding. 
2D MediaPipe poses are first uplifted to constrained 3D skeletons 
using the method of \citet{skeleton_uplift} and then embedded into a 10-dimensional $\beta$-VAE latent space~\cite{higgins2017beta}.
We define \textit{SkeletonVAE Score} as the L2 distance between the reference and hypothesis sequences' DTW-aligned latent trajectories, optionally normalized by the DTW path length.

\paragraph{Skeleton Posterior-based SKL Score}
Following \citet{TarigopulaSGEval2024}, we first extract two sets of linguistically informed features from the pose sequences with the same missing keypoint preprocessing as  Eq. 6 in \citet{arkushin2023ham2pose}. 
For hand movement, we compute 36-dimensional feature vectors representing hand position and velocity relative to the head, shoulders, and hips with a temporal context of 9 frames. 
For handshape, we calculate joint positions relative to the wrist and input them into a separate MLP to obtain handshape posteriors.
The resulting stack of shape and movement posteriors from both the reference 
and hypothesis 
examples is then aligned using DTW with a cost function based on the Symmetric Kullback–Leibler (SKL) divergence.
The cost is aggregated over the DTW time steps as the final score with two variants--\textit{SKL\_mvt Score} (movement only) and \textit{SKL\_mvt\_hshp Score} (movement + handshape), respectively.




\subsubsection{SignCLIP Score}
\label{sec:signclip_score}

One step further than \S\ref{sec:sla_matrics}, we follow \textit{CLIPScore} \cite{hessel2021clipscore} and use SignCLIP \cite{jiang-etal-2024-signclip}, a model repurposed for representing sign language poses by multilingual contrastive learning, to derive \textit{SignCLIPScore P-P} (pose-to-pose), based on the dot product of the embeddings of the reference and hypothesis on the example level instead of frame-level latents plus DTW alignment.

\paragraph{Reference-Free Quality Estimation Variant}
We introduce \textit{SignCLIPScore P-T} (pose-to-text). 
It computes the dot product between the text and pose embedding, eliminating reliance on scarce or even unreliable ground-truth signing references \cite{freitag-etal-2023-results}.



\subsection{Back-Translation-Based Metrics}
\label{sec:bt}

Assuming the existence of the corresponding spoken language text and a reliable pose-to-text SLT model, we can evaluate a sign language pose by:
(a) \textit{Sampling}: translate the pose sequence into text, then compare with the source text using BLEU\footnote{\texttt{nrefs:1|case:mixed|eff:yes|tok:13a|smooth:exp| version:2.3.1}}, chrF\footnote{\texttt{nrefs:1|case:mixed|eff:yes|nc:6|nw:0|space:no| version:2.3.1}}, or BLEURT.
(b) \textit{Scoring}: compute the log-likelihood of the text given the pose sequence as input to the SLT model. This avoids errors introduced by decoding and supports more consistent comparisons across systems.
In this study, we adopt an SLT model from \citet{zhang2024scaling}, which is pretrained on a large-scale YouTube SLT corpus and massive MT data. 
We use \textit{system 8} from their study (\textit{YT-Full + Aug-YT-ASL\&MT-Large + ByT5 XL}), i.e., the current state of the art, and preprocess the generated pose sequences by selecting the same 85 keypoints specified in their paper\footnote{Eight mismatched keypoints due to different MediaPipe versions are imputed as missing landmarks.}.





\section{Automatic Meta-Evaluation}
\label{sec:automatic}

In this section, we explore methods to automatically (meta-)evaluate proposed metrics, especially when there are many variants as seen in \S\ref{sec:distance}.

We adopt the retrieval-based evaluation protocol from \citet{arkushin2023ham2pose} to assess how well different metrics capture meaningful distinctions between signs. Each pose sequence is treated as a query, and the goal is to retrieve other samples of the same sign (\textit{targets}) from a pool that includes unrelated signs (\textit{distractors}).
We focus primarily on the distance-based metric variants introduced in \S\ref{sec:distance}, and compare them against embedding-based alternatives such as \textit{SignCLIP Score} (\S\ref{sec:signclip_score}).

\begin{table*}[ht]
    \centering
    \resizebox{\linewidth}{!}{%
    \begin{tabular}{llccllrr}
\toprule
Base (f) & Fill (e) & Trim (b) &  Norm. (a) & Padding (f) & Keypoints (c) & \textbf{mAP}$\uparrow$ & \textbf{P@10}$\uparrow$ \\
\midrule
Ham2Pose nAPE & 0* &\textcolor{red}{\ding{55}}&\textcolor{green}{\ding{52}}&zero&Reduced&	26\% &	14\%	\\
Ham2Pose nDTW(-MJE) & unspecified &\textcolor{red}{\ding{55}}&\textcolor{green}{\ding{52}}&/&Reduced&	27\% &	14\%	\\
\midrule
APE & 10 & \textcolor{red}{\ding{55}} & \textcolor{red}{\ding{55}} & zero & Upper body &	33\% &	27\%	\\
APE & 10 & \textcolor{red}{\ding{55}} & \textcolor{red}{\ding{55}} & first-frame & Upper body &	34\% &	29\%	\\

APE & 10 & \textcolor{green}{\ding{52}} & \textcolor{red}{\ding{55}} & zero & Upper body &	35\% &	30\%	\\

APE & 10 & \textcolor{red}{\ding{55}} & \textcolor{green}{\ding{52}} & zero & Reduced &	36\% &	32\%	\\

APE & 10 & \textcolor{red}{\ding{55}} & \textcolor{red}{\ding{55}} & first-frame & Reduced &	37\% &	32\%	\\

APE & 10 & \textcolor{red}{\ding{55}} & \textcolor{red}{\ding{55}} & first-frame & YT-ASL &	39\% &	36\%	\\
APE & 10 & \textcolor{green}{\ding{52}} & \textcolor{red}{\ding{55}} & first-frame & Reduced &	40\% &	36\%	\\

APE & 10 & \textcolor{green}{\ding{52}} & \textcolor{green}{\ding{52}} & zero & Upper body &	41\% &	37\%	\\

APE & 10 & \textcolor{red}{\ding{55}} & \textcolor{red}{\ding{55}} & zero & Hands &	42\% &	38\%	\\

APE & 10 & \textcolor{green}{\ding{52}} & \textcolor{green}{\ding{52}} & first-frame & YT-ASL &	43\% &	39\%	\\

APE & 10 & \textcolor{green}{\ding{52}} & \textcolor{red}{\ding{55}} & zero & Hands &	45\% &	41\%	\\

\midrule
DTW & 10 & \textcolor{red}{\ding{55}} & \textcolor{red}{\ding{55}} & / & Upper body &	36\% &	32\%	\\
DTW & 10 & \textcolor{green}{\ding{52}} & \textcolor{red}{\ding{55}} & / & Upper body &	37\% &	33\%	\\
DTW & 10 & \textcolor{red}{\ding{55}} & \textcolor{red}{\ding{55}} & / & Reduced &	42\% &	40\%	\\
DTW & 10 & \textcolor{red}{\ding{55}} & \textcolor{green}{\ding{52}} & / & Upper body &	43\% &	41\%	\\
DTW & 10 & \textcolor{green}{\ding{52}} & \textcolor{green}{\ding{52}} & / & Upper body &	43\% &	41\%	\\

DTW & 10 & \textcolor{red}{\ding{55}} & \textcolor{green}{\ding{52}} & / & Reduced &	44\% &	41\%	\\

DTW & 10 & \textcolor{red}{\ding{55}} & \textcolor{green}{\ding{52}} & / & Hands &	45\% &	41\%	\\

DTW & 10 & \textcolor{red}{\ding{55}} & \textcolor{red}{\ding{55}} & / & YT-ASL &	48\% &	47\%	\\
DTW & 10 & \textcolor{green}{\ding{52}} & \textcolor{red}{\ding{55}} & / & YT-ASL &	49\% &	48\%	\\
DTW & 10 & \textcolor{red}{\ding{55}} & \textcolor{red}{\ding{55}} & / & Hands &	53\% &	52\%	\\
DTW$^{\ddagger}$ & 10 & \textcolor{green}{\ding{52}} & \textcolor{red}{\ding{55}} & / & Hands &	53\% &	52\%	\\
DTW$^{\dagger}$ & 1 & \textcolor{red}{\ding{55}} & \textcolor{green}{\ding{52}} & / & Hands &	55\% &	53\%	\\

\midrule

\multicolumn{6}{l}{SignCLIPScore P-P (multilingual)} &	50\% & 48\% \\

\multicolumn{6}{l}{SignCLIPScore P-P (ASL finetuned)} &	\textbf{91\%} & \textbf{92\%} \\
\bottomrule
\end{tabular}
    }
    \caption{Automatic meta-evaluation of reference-based evaluation metrics on sign retrieval across various settings: (a)-(f) enumerated in section \ref{sec:distance}. 
    The table presents a representative subset of top-performing metrics.
    \textit{Fill} indicates the value used to fill in missing keypoint; \textit{zero} indicates zero-padding; \textit{first-frame} indicates padding with the first frame. \textit{YT-ASL} includes a subset of keypoints used and described in \citet{uthus_youtube-asl_2023}. \textit{Reduced} includes a subset of keypoints used and described in \citet{jiang-etal-2024-signclip}. * Ham2Pose nAPE implements missing-filling slightly differently--filling in zeros for both trajectories if one of them has a missing value (see details in Appendix \ref{app:ham2posedetails}).}
    \label{tab:auto_eval}
\end{table*}


Evaluation is conducted on a combined ASL dataset of ASL Citizen \cite{Desai2023ASLCA}, Sem-Lex \cite{Kezar2023TheSB}, and PopSign ASL \cite{Starner2023PopSignAV}. For each sign/gloss, we use all available samples as targets and sample four times as many distractors, yielding a 1:4 target-to-distractor ratio. For instance, for the sign \textit{HOUSE} with \num{40} samples (\num{11} from ASL Citizen, \num{29} from Sem-Lex), we add \num{160} distractors and compute pairwise distance from each target to all \num{199} other examples.
The pairwise distance is defined by each of these proposed metric scores.
Retrieval quality is measured using Mean Average Precision (mAP$\uparrow$) and Precision@10 (P@10$\uparrow$). The full evaluation covers \num{5362} unique signs and \num{82099} pose sequences. 
After several pilot runs to rule out clear bad choices, we finalize a subset of \num{169} signs with at most \num{20} samples each,
and evaluate \num{48} representative keypoint distance-based metric candidates and \textit{SignCLIP Score} with different SignCLIP checkpoints provided by the authors\footnote{\url{https://github.com/J22Melody/fairseq/tree/main/examples/MMPT\#demo-and-model-weights}} on this subset. 
For reference, we also reproduce the metrics proposed by \citet{arkushin2023ham2pose}.
The key results, including the best metrics, are presented in Table~\ref{tab:auto_eval}.

The results show that, as expected, DTW-based metrics outperform padding-based APE baselines. 
While selecting hands-only keypoints appears to yield the best results, a more sophisticated selection that includes non-manuals might still be desirable.
Embedding-based methods, particularly SignCLIP models fine-tuned on in-domain ASL data, achieve the strongest retrieval scores.
We mark the two best DTW-based metrics by $^{\ddagger}$ and $^{\dagger}$, and rename them \textit{DTW$p$} and \textit{nDTW$p$} for use in the rest of the paper.

\section{Text-to-Pose Translation Study with Human Evaluation}
\label{sec:correlation}

This section shifts our evaluation focus from automatic sign-level tasks to a sentence-level text-to-pose sign language machine translation scenario.
Due to their subjective and diverse nature, open-ended text or utterance generation tasks inherently lack a single ``correct''/``ground-truth'' answer. 
Consequently, \textbf{automatic evaluation metrics are only \textit{meaningful} if they correlate closely with human judgments} \cite{reiter-2018-structured,sellam-etal-2020-bleurt}.





\begin{table*}
    \centering
    \resizebox{\linewidth}{!}{%
    \begin{tabular}{l rrrr rrrr rr rrrr r}
      \toprule
      \multicolumn{1}{c}{} &
      \multicolumn{9}{c}{\textbf{Reference‐Based}} &
      \multicolumn{6}{c}{\textbf{Reference‐Free}}  \\
      \cmidrule(lr){2-10} \cmidrule(lr){11-16}
      \multicolumn{1}{c}{} &
      \multicolumn{4}{c}{\textbf{Distance‐Based}} &
      \multicolumn{4}{c}{\textbf{SLA Metrics}} &
      \multicolumn{2}{c}{\textbf{SignCLIPScore}} &
      \multicolumn{4}{c}{\textbf{Back Translation‐Based}} &
      \textbf{H*}  \\
      \cmidrule(lr){2-5} \cmidrule(lr){6-9} \cmidrule(lr){10-11} \cmidrule(lr){12-15} \cmidrule(lr){16-16}
      & \textbf{nAPE}
      & \textbf{nDTW}
      & \textbf{DTW$p$}
      & \textbf{nDTW$p$}
      & \textbf{SVAE}
      & \textbf{SVAE$_n$}
      & \textbf{SKL}
      & \textbf{SKL$_h$}
      & \textbf{P‐P}
      & \textbf{P‐T}
      & \textbf{B4}
      & \textbf{chrF}
      & \textbf{B‐RT}
      & \textbf{Lik.}
      & \textbf{H*}  \\
      \midrule
      \multicolumn{15}{l}{\textit{By System}}  \\
      \noalign{\vskip 1pt}
      \href{https://sign.mt}{sign.mt}       &  0.09 &  0.14 &  0.11 &  0.10 &  0.23 & -0.08 &  0.24 & 0.17 &  0.10 &  0.02 &  0.05 &  0.11 &  0.05 &  0.23 &  0.43  \\
      \href{https://sign.mt}{sign.mt} v2   &  0.28 &  0.33 &  0.26 &  0.31 &  0.46 &  0.14 &  0.22 & 0.29 &  0.00 & -0.19 &  0.20 &  0.22 &  0.44 &  0.49 &  0.52  \\
      Sockeye        &  0.10 &  0.15 &  0.04 &  0.17 &  0.13 &  0.01 &  0.24 & 0.07 &  0.42 & -0.27 & -0.07 &  0.04 &  0.46 &  0.58 &  0.22  \\
      \midrule
      \multicolumn{15}{l}{\textit{By Language}}  \\
      \noalign{\vskip 1pt}
      DE$\to$DSGS    & -0.36 & -0.09 &  0.73 &  0.43 & -0.02 &  0.27 & -0.57 & -0.51 & -0.31 &  0.39 &  0.18 &  0.26 &  0.09 &  0.36 &  0.70  \\
      FR$\to$LSF     & -0.54 & -0.11 &  0.76 &  0.02 & -0.01 &  0.37 & -0.68 & -0.65 & -0.01 &  0.45 &  0.32 &  0.60 &  0.47 &  0.29 &  0.80  \\
      IT$\to$LIS     & -0.57 & -0.39 &  0.79 &  0.57 & -0.02 &  0.53 & -0.75 & -0.74 &  0.13 &  0.29 &  0.31 &  0.63 &  0.41 &  0.38 &  0.88  \\
      \midrule
      Overall ($\uparrow$) & -0.41 & -0.10 &  0.76 &  0.43 &  0.07 &  0.38 & -0.56 & -0.53 & -0.10 &  0.27 &  0.21 &  0.42 &  0.36 &  0.42 &  0.77  \\
      {\footnotesize SD ($\downarrow$)}
                       & {\footnotesize(0.35)} & {\footnotesize(0.24)} & {\footnotesize(0.34)} & {\footnotesize(0.20)} & {\footnotesize(0.18)} & {\footnotesize(0.22)} & {\footnotesize(0.47)} &
                       {\footnotesize(0.43)} &{\footnotesize(0.22)} & {\footnotesize(0.29)} & {\footnotesize(0.14)} & {\footnotesize(0.23)} & {\footnotesize(0.18)} & {\footnotesize(0.12)} & {\footnotesize(0.24)}  \\
      \bottomrule
    \end{tabular}%
    }
    \caption{Segment‐level Spearman correlations with average human judgments calculated for several pose‐based evaluation metrics for sign language.  
    nAPE=\texttt{normalized APE}, nDTW=\texttt{normalized DTW‐MJE} (two metrics taken from \citet{arkushin2023ham2pose} and re-implemented for MediaPipe, normalized by pose shoulder);    
    DTW$p$=\texttt{DTW+Trim+MaskFill10.0+Hands-Only}, nDTW$p$=\texttt{DTW+MaskFill1.0+Norm.+Hands-Only} (top metrics selected in \S\ref{sec:automatic} implemented by \texttt{pose-evaluation}, denoted by $^{\ddagger}$ and $^{\dagger}$ in Table \ref{tab:auto_eval}, without/with pose normalization, respectively);  
    SVAE=\texttt{SkeletonVAE Score}, SVAE$_n$=\texttt{SVAE normalized by DTW path},  
    SKL=\texttt{SKL\_mvt Score}, SKL$_h$=\texttt{SKL\_mvt\_hshp Score};  
    P‐P=\texttt{Pose‐to‐pose embedding distance}, P‐T=\texttt{Pose‐to‐text embedding distance};  
    B4=\texttt{BLEU‐4}, chrF=\texttt{chrF}, B‐RT=\texttt{BLEURT}, Lik.\space=\texttt{Likelihood}.  
    H* denotes mean inter‐evaluator Spearman correlation. SD represents the standard deviation across each column and is expected to be small for an ideal metric.}
    \label{tab:correlation}
\end{table*}

\subsection{Dataset: WMT-SLT Signsuisse}

We use the \href{https://www.sgb-fss.ch/signsuisse/}{Signsuisse} dataset released in the \href{https://www.wmt-slt.com/data\#h.l0qcgunwhkqt}{WMT-SLT 23} campaign.
The dataset comprises 18,221 lexical items in three spoken-sign language pairs, represented as videos and glosses.
One signed example sentence for each lexical item is presented in a video along with the corresponding spoken language translation, which forms parallel data between the sign and spoken languages. 
The test set is used to test different text-to-pose translation systems. It contains 500 German/Swiss German Sign Language (DSGS) segments, 250 French/French Sign Language (LSF) segments, and 250 Italian/Italian Sign Language (LIS) segments.

\subsection{Systems}

We utilize three text-to-pose translation systems that convert spoken language text inputs into corresponding sign language represented by the MediaPipe Holistic pose formats.

\paragraph{Reference*} MediaPipe poses are estimated from the reference translation videos, i.e., ground truth.

\paragraph{\href{https://sign.mt}{sign.mt}} 
Based on \citet{moryossef-etal-2023-open}, this open system converts text into sign language glosses through rule-based reordering and selective word dropping. Glosses are mapped to skeletal poses retrieved from a lexicon and are then concatenated to form coherent sequences. When a gloss is missing from the lexicon, the system defaults to fingerspelling the corresponding word.

\paragraph{\href{https://sign.mt}{sign.mt} v2} 
During evaluation, we found that frequent fingerspelling of missing glosses was cumbersome and frustrating for evaluators. Therefore, in this version, we opted to omit glosses without lexical mappings, acknowledging that while this may result in information loss, it significantly improves user experience and evaluation efficiency.

\paragraph{Sockeye}
We adapt Sockeye \cite{hieber2022sockeye} to continuous pose sequences by modifying both the encoder and decoder to handle continuous sequences.
The text-to-pose Sockeye model is trained on the Signsuisse training set with 60k updates on a 32GB NVIDIA Tesla V100 GPU.





To avoid exposure bias—where the decoder overfits to gold frames and fails at inference, we first predict only the initial pose $y_1$ from the encoder output, then feed $y_1$ as input for all subsequent steps $y_{2:n}$, training the decoder to output frame‐to‐frame deltas $\Delta y_t = y_t - y_1$ instead of absolute poses.
Since the target sequence is continuous, we replace the cross-entropy loss function with mean squared error on the poses. Additionally, there is no \texttt{<EOS>} token with continuous output; instead, we learn to output the length of the pose sequence based on the length ratios from the training data.
We provide the link to the adapted \href{https://github.com/ZurichNLP/sockeye/tree/continuous_outputs_3.2}{Sockeye repository} and a \href{https://github.com/ZurichNLP/sockeye/blob/continuous_outputs_3.2/pose_examples_signsuisse/README.md}{demo} of translation output.

\subsection{Human Evaluation}

We collect system translations and use \href{https://github.com/J22Melody/Appraise/tree/iict}{Appraise} (\citet{federmann-2018-appraise}; Figure \ref{fig:appraise_screenshot}) to allow evaluators to rate the translations on a continuous scale between 0 and 100 as in traditional direct assessment \cite{graham-etal-2013-continuous,cettolo2017overview} but with 0-6 markings on the analogue slider and custom annotator guidelines designed explicitly for our task (similar to WMT-SLT, but reverse translation direction).
Evaluation instructions are sent out in 
DSGS, LSF, and LIS, which are translations of the respective spoken language instructions in WMT-SLT.
The instructions are attached in Appendix \ref{app:huamn-eval-details}.

We hire seven DSGS, two LSF, and four LIS evaluators, all of whom are native deaf sign language users\footnote{One additional DSGS evaluator, a hearing interpreter, did a pilot study with us to test the Appraise system.}. 
All work is done with informed consent in written and signed form.
Of the seven native DSGS deaf signers, four have never participated in such an evaluation campaign before, two have participated once, and one has attended more than once. Concerning their professional backgrounds, four are deaf translators; one also interprets live.  
Complete demographics are presented in Table \ref{tab:raterdemographics}.

An initial round of evaluation informs us about the cost, roughly 100 example segments per hour, with a compensation of $\sim$40 USD per hour. 
Evaluators also provide constructive feedback on the Appraise platform and the translation systems, which results in the switching into the v2 version of \href{https://sign.mt}{sign.mt}. Therefore, the number of evaluated examples varies slightly between systems and languages.

\paragraph{Statistics}

The evaluation comprises \num{2650} unique examples and \num{11471} ratings across all four systems (\num{3275} reference, \num{4032} Sockeye, \num{861} sign.mt, and \num{3303} sign.mt v2) and three language pairs (\num{7861} DSGS, \num{1210} LSF, and \num{2400} LIS).

We follow the practices set by WMT-SLT.
The inter-annotator agreement, measured with an approximation of Fleiss $\kappa$ \cite{Fleiss1971} by discretizing the continuous scale 0-100 in seven bins in the scale 0-6, is $\kappa=0.36\pm0.05$.
We also randomly mix 500 references and some repeated hypothesis segments for sanity checks and quality control.
The mean intra-annotator agreement over all evaluators is $\kappa=0.49\pm0.09$, calculated over 50-100 segments evaluated twice by the same evaluator.
We find the inter- and intra-annotator agreement to be lower than in the WMT-SLT study for the sign-to-text translation direction, and posit that the lack of a clear definition and criteria for translation quality in sign language poses a significant challenge.

\begin{table}[ht]
  \centering
  \resizebox{\linewidth}{!}{%
  \begin{tabular}{l rrr rrr r}
    \toprule
    & \multicolumn{3}{c}{Evaluation experience} 
    & \multicolumn{3}{c}{SL professional} 
    & Avg yr. \\
    \cmidrule(lr){2-4} \cmidrule(lr){5-7}
    & Never & Once & > Once 
    & Translator & Interpreter & Teacher 
    & \\
    \midrule
    DSGS (7) & 4 & 2 & 1 & 4 & 1 & 4 & 39.0 \\
    LSF  (2) & 0 & 1 & 1 & 1 & 0 & 1 & 35.0 \\
    LIS  (4) & 1 & 1 & 2 & 3 & 4 & 0 & 42.5 \\
    \bottomrule
  \end{tabular}
  }
  \caption{Raters overview: system evaluation and professional experience with sign language, average number of years signing (in most cases equivalent to age).}
  \label{tab:raterdemographics}
\end{table}

\subsection{Correlation Analysis}

We perform a correlation analysis between the metrics proposed in \S\ref{sec:metrics} and the human scores averaged over evaluators at the segment level, as presented in Table \ref{tab:correlation}.
The metrics are divided into families, where \textit{reference-based} means that the quality of the translated poses is measured in relation to reference poses derived from signing videos. 
The absolute scores per metric/system are presented in Table \ref{tab:absolute_scores}.

\begin{table*}[ht]
  \centering
  \resizebox{\linewidth}{!}{%
    \begin{tabular}{l r r r r r r r r r r r r r r r}
      \toprule
       & \textbf{nAPE}$\downarrow$
       & \textbf{nDTW}$\downarrow$
       & \textbf{DTW$p$}$\downarrow$
       & \textbf{nDTW$p$}$\downarrow$
       & \textbf{SVAE}$\downarrow$
       & \textbf{SVAE$_n$}$\downarrow$
       & \textbf{SKL}$\downarrow$
       & \textbf{SKL$_h$}$\downarrow$
       & \textbf{P-P}$\uparrow$
       & \textbf{P-T}$\uparrow$
       & \textbf{B4}$\uparrow$
       & \textbf{chrF}$\uparrow$
       & \textbf{B-RT}$\uparrow$
       & \textbf{Lik.}$\uparrow$
       & \textbf{H*}  \\
      \midrule
      reference*   & n/a   & n/a   & n/a   & n/a   & n/a    & n/a   & n/a    & n/a     & n/a    & 74.23  & 15.05 & 38.52 & 0.49  & -32.87 & 76.55 \\
      \href{https://sign.mt}{sign.mt}     & 0.60  & 25.43 & 5171.55  & 8.66   & 1.20  & 0.0028 & 2794.29 & 7443.66 & 81.58  & 75.55  & 3.46  & 14.53 & 0.26  & -39.30 & 22.00 \\
      \href{https://sign.mt}{sign.mt} v2 & 1.65  & 20.73 & 4508.62  & 8.97   & 1.23  & 0.0045 & 4165.78 & 11540.15 & 89.89  & 76.46  & 6.89  & 24.79 & 0.23  & -67.55 & 30.12 \\
      Sockeye      & 0.29  & 16.97 & 11879.58 & 10.71  & 1.16  & 0.0057 & 1056.29 & 3985.65 & 94.45  & 72.40  & 3.44  & 11.90 & 0.17  & -77.57 &  5.05 \\
      \bottomrule
    \end{tabular}%
  }
  \caption{Mean absolute scores for each metric across systems. Rows and columns mirror those in Table~\ref{tab:correlation}.}
  \label{tab:absolute_scores}
\end{table*}

For the distance-based metrics, we reproduce \textit{nAPE} and \textit{nDTW(-MJE)} for MediaPipe poses based on the open implementation from \citet{arkushin2023ham2pose} as a reference, and additionally compare them to the best-performing metrics informed by the automatic meta-evaluation in \S\ref{sec:automatic} on ASL, a different sign language.
We flip the signs of the metrics that quantify errors to keep a positive correlation for analytical convenience.
Row-wise, we first break down the correlation by system and language into relevant rows, and then present the overall correlation, including all systems and languages, to reflect performance at the system level.

\section{Discussion and Recommendations}
\label{sec:discussion}

\paragraph{Distance-based metrics are efficient defaults, but the devil is in the implementation details.}
Although seemingly straightforward to implement, distance-based metrics involve many design choices, including pose format and keypoint selection.
We empirically demonstrate the effectiveness of correcting these choices through a random parameter search, following our established meta-evaluation protocols in \S\ref{sec:automatic}.
We recommend using the tuned versions—\textit{DTW$p$} and \textit{nDTW$p$}—in our \texttt{pose-evaluation} library, or tuning your distance-based metrics when necessary.

\paragraph{Upon successful tuning, a distance-based metric achieves decent sign retrieval and correlation with humans in text-to-pose translation.}
Our tuned metrics can even be used as a distance function for a nearest neighbor classifier\footnote{Upon quick experimentation, \textit{nDTW$p$} with KNN (n=10) achieves 19\% ISLR accuracy on the ASL Citizen test set. We leave a more systematic evaluation on this end to future work.}, and reach close performance as the SignCLIP model pretrained on multilingual sign language data; still, it lags behind a SignCLIP model fine-tuned on in-domain data (Table \ref{tab:auto_eval}).
When used to evaluate translation output, keypoint distance-based metrics can range from negatively correlated with human judgments (as seen for \textit{nAPE} and \textit{nDTW}), to being the best metrics tested. 
\textit{DTW$p$} wins the overall correlation while \textit{nDTW$p$} is more sensitive on the segment level within a specific system (Table \ref{tab:correlation}).

\paragraph{SLA metrics correlate with humans on the segment level but are confused on the system level.}
While verified to align with human ratings for their tasks on evaluating human-produced signing (usually involving fixed individual signs), text-to-pose translation is more lengthy and open-ended, which hinders direct transferability. A proper length normalization (as seen in the case of \textit{SVAE$_n$} vs. \textit{SVAE}) might help on the system level at the price of losing precision on the segment level.

\paragraph{SignCLIP, used as a multilingual embedding device, excels on the sign level, but falls short for sentence-level translation evaluation.} 
We speculate that using a single embedding to summarize a long-duration ($>$ 10 seconds) signing video is inherently limited, especially for DSGS, a language unseen during SignCLIP's pretraining. Nevertheless, the reference-free variant exhibits a moderate correlation at the system level, and we observe a similar tradeoff (\textit{P-P} vs. \textit{P-T}) between segment and system level correlations, as seen in the SLA metrics.

\paragraph{Back-translation-based approaches correlate properly with human judgment; a gap remains compared to inter-human correlation.}
In addition to the standard practices suggested by \citet{muller-etal-2023-considerations} on computing text-based metrics, we call for open, standardized pose-to-text translation models that include both the model weights and the source code.
Yet, as noted in Table \ref{tab:literature_review}, this is hardly the case in current research, and having a dedicated back-translation model for each translation direction (or even dataset) is a luxury.
The above-mentioned metrics, which do not rely on in-domain data but function to a decent degree, are valuable in a more generic setting.
Human evaluation shall be used as the final quality assessment resort.

\paragraph{When using back translation, likelihood is consistent and more reliable than text metrics.}
BLEU, chrF, and BLEURT show weaker or unstable correlations with humans in Table \ref{tab:correlation}. 
It is recommended that back-translation likelihood be included as a primary metric when a pose-to-text model is available.


\section{Conclusion}

This work presents a unified framework and an open‐source \texttt{pose-evaluation} toolkit for systematically assessing (generated) sign language utterances based on human skeletal poses. 
We implemented and compared a wide range of metrics (\S\ref{sec:metrics})—distance-based, embedding-based, and back-translation-based—via automatic meta-evaluation on sign retrieval (\S\ref{sec:automatic}) and a comprehensive human correlation study across three sign languages (\S\ref{sec:correlation}). 
Our results demonstrate that carefully tuned distance metrics, namely \textit{DTW$p$} and \textit{nDTW$p$}, and back-translation likelihoods yield the strongest agreement with native signer judgments. 
We release our code, evaluation protocols, and human ratings to foster reproducible and fair comparisons in computational sign language research. 

\section{Limitations}
\label{sec:limitation}

\subsection{3D Pose Representation}

While our study focuses on using MediaPipe Holistic as the pose format for representing sign language motion, other specifics, especially the recently developed 3D SMPL-X \cite{Pavlakos_2019_CVPR} would be a visually more expressive choice.
However, the lack of a common way to extract and use 3D poses as easily as MediaPipe Holistic makes the latter the most used choice in SLP.

\subsection{Missing Publicly Available Systems}

Our study is further limited by the number of public systems (Table \ref{tab:literature_review}) we can use to run the correlation analysis, unless we implement everything from scratch (including the pose estimation pipelines, text-to-pose systems, and back-translation models).
We hope the release of this work will alleviate the situation.

\subsection{Automatic Evaluation beyond Sign Level}

The automatic meta-evaluation in \S\ref{sec:automatic} is capped by the sign-level retrieval task, and we envision extending it to the phrase level.
One possible approach is to leverage the Platonic Representation Hypothesis proposed by \citet{pmlr-v235-huh24a}.
In the pose evaluation scenario, we hypothesize that the similarity given by a good pose metric between two pose segments should correlate with the similarity given by a text embedding model between the two text segments paired with the two pose segments, respectively.
We leave exploration on this end for future work, which will likely connect the automatic meta-evaluation more closely to the sentence-level human correlation study in \S\ref{sec:correlation}.

\subsection{Tokenized Evaluation}

Inspired by how text metrics like BLEU collect surface-form overlapping statistics, we envision a tokenized evaluation as promising for sign language evaluation. 
Although a sign language pose sequence cannot be discretely tokenized and matched like text tokens, the combination of a sign language segmentation model \cite{moryossef-etal-2023-linguistically} plus SignCLIP embedding can be utilized in a way similar to BERTScore \cite{bert-score}, where a similarity matrix is constructed between the reference and hypothesis tokens to derive the final similarity score on phrase level.

\section*{Acknowledgments}

This work is funded by the Swiss Innovation Agency (Innosuisse) flagship IICT (PFFS-21-47) and by the SIGMA project at the UZH Digital Society Initiative (DSI).

We thank the deaf evaluators for their efforts in the human evaluation and valuable feedback on the evaluated systems.
We thank SWISS TXT for helping organize the evaluation campaign.
We thank Roman Grundkiewicz for troubleshooting the Appraise platform and thank Lisa Arter for a pilot test on it.
We thank Andreas Säuberli for the insightful discussion on inter-annotator agreement.
We also thank Garrett Tanzer for answering questions about the MediaPipe version used in Google's pose-to-text translation system, and Ronglai Zuo for the advice on the paper draft.



\bibliography{custom}

\newpage
\clearpage
\appendix

\section{Ham2Pose Metrics Re-Implementation via \texttt{pose-evaluation}}
\label{app:ham2posedetails}
\texttt{pose-evaluation} toolkit enables the flexible creation of various metrics and pose processing pipelines. 
We successfully re-implemented the \textit{nMSE}, \textit{nAPE}, and \textit{nDTW-MJE} metrics, and verified that the new implementation gave exactly identical results on a small collection of test files. 

All metrics share certain preprocessing steps before comparison, in this order:

\begin{enumerate}
    \item Remove \texttt{world landmarks}.
    \item Calling the \verb|reduce_holistic| function from the \href{https://github.com/sign-language-processing/pose}{\texttt{pose-format}} library, effectively reducing the keypoints to the face contour and the upper body.
    \item Normalization by shoulder joints.
    \item Hide low-confidence joint predictions.
\end{enumerate}

In addition, \textit{nMSE} and \textit{nAPE} metrics do trajectory-based preprocessing, for each pair of keypoint trajectories:
\begin{enumerate}
    \item Zero-pad the shorter trajectory. 
    \item Fill with zeros anywhere where either one of the trajectories is missing a value. For example, if trajectory A had [7, ---, 7] and trajectory B had [---,8,8], the result would be thus: Trajectory A: [0,0,7], B: [0,0,8]. 
\end{enumerate}

In contrast, the metrics implemented for the automatic meta-evaluation in \S\ref{sec:automatic}, e.g., \texttt{DTW+Trim+MaskFill10.0+Hands-Only}, behave differently, filling each pose in without regard to the other. The result of trajectory A = [7,---,7] vs  Trajectory B [---,8,8] would thus become  A = [7,10,7] vs  Trajectory B [10,8,8]. 



\section{Extended Human Evaluation Details}
\label{app:huamn-eval-details}

\begin{figure*}[ht]
\begin{center}
\includegraphics[width=\textwidth]{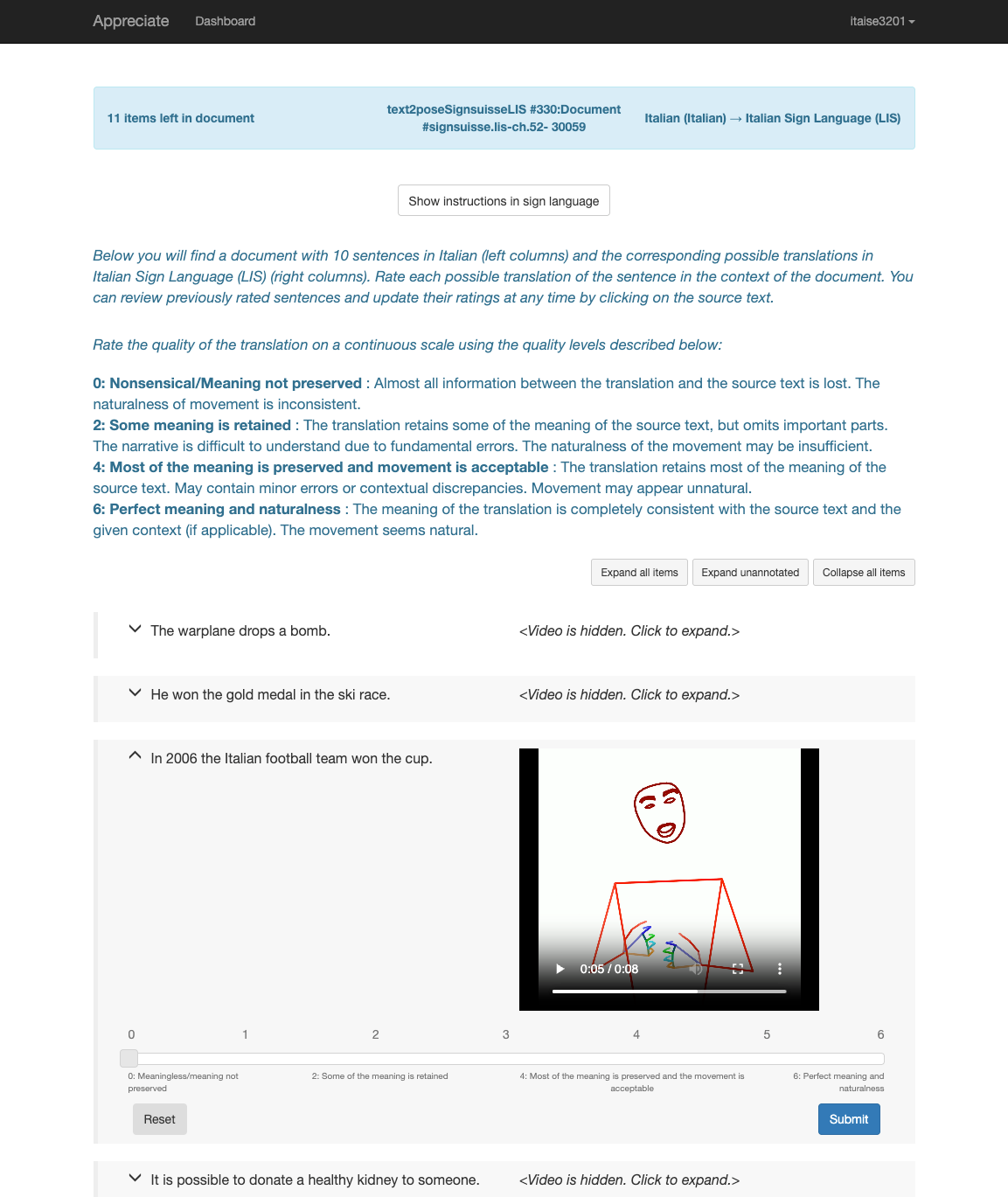}
\caption{A screenshot of an example text-to-pose evaluation task in Appraise featuring sentence-level source-based direct assessment with custom annotator guidelines in German/French/Italian and DSGS/LSF/LIS, translated into English for readers' convenience.}
\label{fig:appraise_screenshot}
\end{center}
\end{figure*}

Figure \ref{fig:appraise_screenshot} presents a screenshot of the Appraise platform we customized for the text-to-pose evaluation, where the instruction text is translated into English.
The sign language versions of the instructions are linked: \href{https://uzh.mediaspace.cast.switch.ch/embed/secure/iframe/entryId/0_ginojjwt/uiConfId/23448425/st/0}{DSGS}, \href{https://uzh.mediaspace.cast.switch.ch/embed/secure/iframe/entryId/0_q1uhudq3/uiConfId/23448425/st/0}{LSF}, \href{https://uzh.mediaspace.cast.switch.ch/embed/secure/iframe/entryId/0_m8sapvzb/uiConfId/23448425/st/0}{LIS}.
The original text instructions in German, French, and Italian are below:

\paragraph{German} 
Unten sehen Sie 10 Sätzen auf Deutsch (linke Spalten) und die entsprechenden möglichen Übersetzungen in Deutschschweizer Gebärdensprache (DSGS) (rechte Spalten). Bewerten Sie jede mögliche Übersetzung des Satzes. Sie können bereits bewertete Sätze jederzeit durch Anklicken eines Quelltextes erneut aufrufen und die Bewertung aktualisieren.

Bewerten Sie die Übersetzungsqualität auf einer kontinuierlichen Skala mit Hilfe der nachfolgend beschriebenen Qualitätsstufen:

0: Unsinn/Bedeutung nicht erhalten: Fast alle Informationen zwischen Übersetzung und Ausgangstext sind verloren gegangen. Es ist irrelevant, ob die Bewegungen natürlich sind.

2: Ein Teil der Bedeutung ist erhalten: Die Übersetzung behält einen Teil der Bedeutung der Quelle bei, lässt aber wichtige Teile aus. Die Erzählung ist aufgrund von grundlegenden Fehlern schwer zu verstehen. Bewegungen können mangelhaft sein.

4: Der grösste Teil der Bedeutung ist erhalten und die Bewegungen sind akzeptabel: Die Übersetzung behält den grössten Teil der Bedeutung der Quelle bei. Sie kann kleine Fehler oder kleinere kontextuelle Unstimmigkeiten aufweisen. Bewegungen sehen teilweise nicht natürlich aus.

6: Perfekte Bedeutung und Natürlichkeit: Die Bedeutung der Übersetzung stimmt vollständig mit der Quelle und dem umgebenden Kontext (falls zutreffend) überein. Bewegungen wirken natürlich.

\paragraph{French}
Vous voyez ci-dessous un document avec 10 phrases en français (colonnes de gauche) et leurs traductions candidates correspondantes langue des signes française (LSF) (colonnes de droite). Veuillez attribuer un score à chaque traduction possible de la phrase dans le contexte du document. Vous pouvez revisiter les phrases déjà évaluées et mettre à jour leurs scores à tout moment en cliquant sur un texte source.

Évaluez la qualité de la traduction sur une échelle continue en utilisant les niveaux de qualité décrits ci-dessous:

0: Absence de sens/aucune signification préservée: Presque toutes les informations sont perdues entre la traduction et la source. Le caractère naturel du mouvement n'est pas pertinent.

2: Une partie du sens est préservée: La traduction préserve une partie du sens de la source mais omet des parties importantes. Le récit est difficile à suivre en raison d'erreurs fondamentales. Le mouvement n'est pas toujours naturel.

4: La majeure partie du sens est préservée et le caractère naturel du mouvement est acceptable: La traduction conserve la majeure partie du sens de la source. Elle peut comporter quelques erreurs mineures ou des incohérences contextuelles. Le mouvement peut sembler peu naturel.

6: Sens parfait et mouvements naturels: Le sens de la traduction est totalement cohérent avec la source et le contexte environnant (le cas échéant). Les mouvements sont naturels.

\paragraph{Italian}
Qui sotto trovate un documento con 10 frasi in italiano (colonne di sinistra) e lecorrispondenti possibili traduzioni nella lingua dei segni italiana (LIS) (colonne di destra). Valutate ogni possibile traduzione della frase nel contesto del documento. Potete rivedere le frasi valutate in precedenza e aggiornarne le valutazioni in qualsiasi momento cliccando sul testo sorgente.

Valutate la qualità della traduzione su una scala continua utilizzando i livelli di qualità descritti di seguito:

0: Privo di senso/significato non conservato: Quasi tutte le informazioni tra la traduzione e il testo sorgente sono andate perse. La naturalezza del movimento è inconsistente.

2: Parte del significato è conservato: La traduzione conserva parte del significato del testo sorgente, ma omette parti importanti. La narrazione è difficile da capire a causa di errori fondamentali. La naturalezza del movimento può essere insufficiente.

4: La maggior parte del significato è conservato e il movimento è accettabile: La traduzione conserva la maggior parte del significato del testo sorgente. Può contenere errori o discrepanze contestuali di entità minore. Il movimento può sembrare innaturale.

6: Significato perfetto e naturalezza: Il significato della traduzione è completamente coerente con il testo sorgente e con il contesto dato (se applicabile). Il movimento sembra naturale.

\end{document}

%% file: figures/title-figure.tex
\begin{tikzpicture}[
  node distance=1cm,
  every node/.style={font=\small, align=center},
  box/.style={
    draw=gray!50,
    line width=0.5pt,
    rectangle,
    rounded corners,
    inner sep=2pt
  }
]
\node (ref) [box, xshift=-2.5cm] {
  \begin{tikzpicture}[x=1cm,y=1cm]
    \node[anchor=west,opacity=0.5] at (0,0)    {\includegraphics[width=2cm,height=2cm]{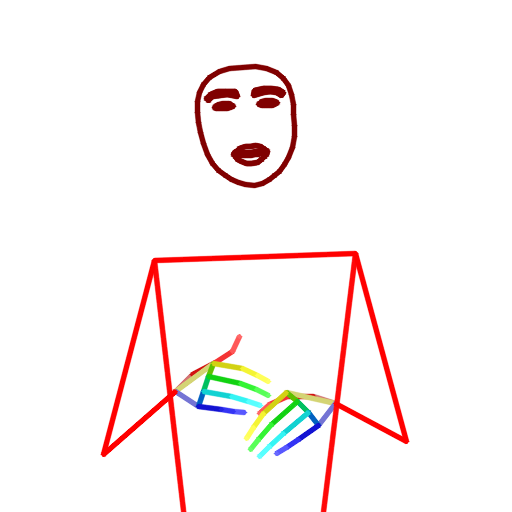}};
    \node[anchor=west,opacity=0.7] at (-0.5,-0.5){\includegraphics[width=2cm,height=2cm]{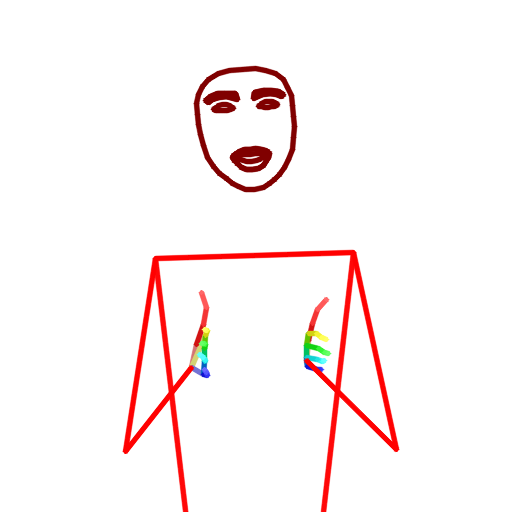}};
    \node[anchor=west]            at (-1,-1)  {\includegraphics[width=2cm,height=2cm]{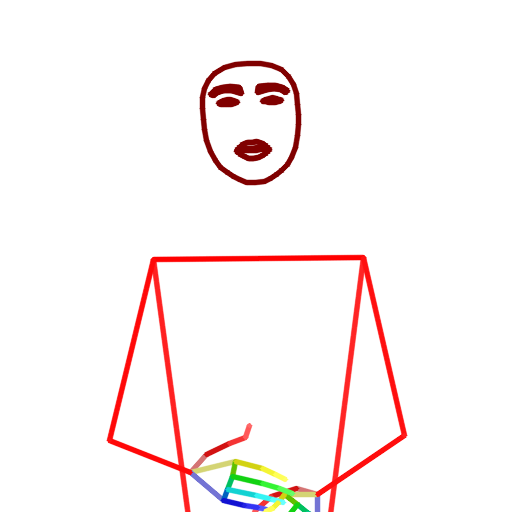}};
  \end{tikzpicture}
};
\node[above=2pt of ref] {Reference};

\node (hyp) [box, xshift=2.5cm] {
  \begin{tikzpicture}[x=1cm,y=1cm]
    \node[anchor=west,opacity=0.5] at (0,0)    {\includegraphics[width=2cm,height=2cm]{figures/poses/kleine/49.png}};
    \node[anchor=west,opacity=0.7] at (-0.5,-0.5){\includegraphics[width=2cm,height=2cm]{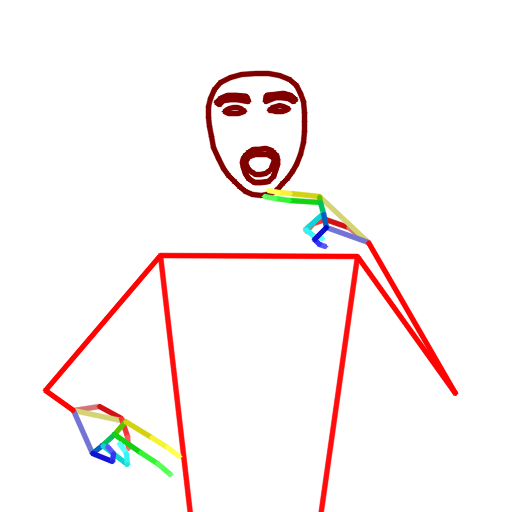}};
    \node[anchor=west]            at (-1,-1)  {\includegraphics[width=2cm,height=2cm]{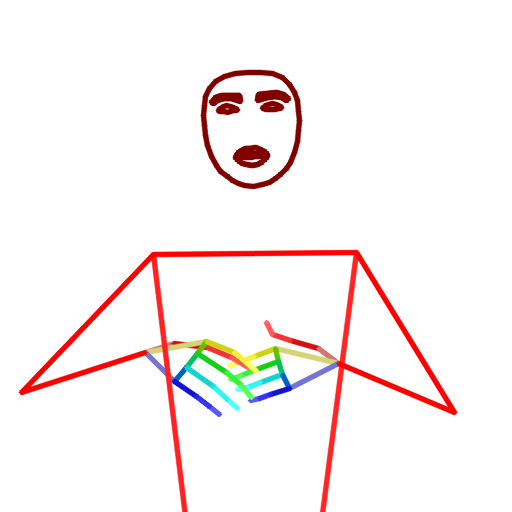}};
  \end{tikzpicture}
};
\node[above=2pt of hyp] {Hypothesis};

\draw[<->, dashed] 
  (ref.east) 
  -- node[above] {distance}
     node[below] {\scriptsize (L1, L2, DTW)} 
  (hyp.west);

\node (emb_ref) [ellipse, fill=green!20, below=of ref, yshift=0.5cm] {Embed};
\node (emb_hyp) [ellipse, fill=green!20, below=of hyp, yshift=0.5cm] {Embed};
\draw[->, thick] (ref.south) -- (emb_ref.north);
\draw[->, thick] (hyp.south) -- (emb_hyp.north);
\draw[<->, dashed] 
  (emb_ref) 
  -- node[above] {vector similarity}
     node[below] {\scriptsize (dot product, cosine similarity)} 
  (emb_hyp);
  
\node (bt) [ellipse, fill=orange!20, below=of emb_hyp, minimum height=1cm, yshift=0.5cm] {Translate};
\draw[->, thick] ([xshift=1.3cm]hyp.south) -- ++(0,0) |- (bt.east);

\node (text) [draw=gray!50, line width=0.5pt, rounded corners, below=of emb_ref, minimum width=2.5cm, minimum height=1cm, yshift=0.5cm] {Text};
\draw[<->, dashed] 
  (text.east) 
  -- node[above] {text similarity}
     node[below] {\scriptsize (BLEU, chrF)} 
  (bt.west);

\draw[-, thick] ([xshift=-1cm]ref.south) -- node[above, rotate=90] {pair} ([xshift=-1cm]text.north);

\end{tikzpicture}